\documentclass[letterpaper]{article}
\usepackage[preprint]{aaai2027}
\usepackage[hyphens]{url}
\usepackage{graphicx}
\usepackage{natbib}
\usepackage{caption}
\usepackage{booktabs}
\usepackage{colortbl}
\usepackage{amsmath}
\usepackage[hidelinks]{hyperref}
\usepackage{etoolbox}

\AfterEndEnvironment{abstract}{%
  \begin{center}
  \small\textbf{Code:} \url{https://github.com/jn12-29/SemPIC}
  \end{center}
}

\title{SemPIC: Learning Semantic Position-Independent KV Caches}
\author{
Hui Xie\textsuperscript{\rm 2,\rm 4},
Peng Xiao\textsuperscript{\rm 3},
Yutong Deng\textsuperscript{\rm 1},
Shuoran Dou\textsuperscript{\rm 1},
Jian Yang\textsuperscript{\rm 1,\rm 4},
Jinyang Guo\textsuperscript{\rm 2,\rm 4}\corresponding
}
\affiliations{
\textsuperscript{\rm 1}School of Computer Science and Engineering, Beihang University\\
\textsuperscript{\rm 2}Institute of Artificial Intelligence, Beihang University\\
\textsuperscript{\rm 3}Shen Yuan Honors College, Beihang University\\
\textsuperscript{\rm 4}State Key Laboratory of Complex \& Critical Software Environment\\
\{xiehui, xiaop, 511170, 18294491955, jiayang, jinyangguo\}@buaa.edu.cn
}

\begin{document}
\maketitle
\begin{abstract}
Long-context retrieval and agentic workloads repeatedly reuse the same documents under changing instructions, histories, and document orders. Prefix caching cannot exploit this reuse, while position-independent caching (PIC) remains unreliable because independently compiled KV states lack the future context in which they will be consumed. Our diagnostics show that a learned boundary-conditioned baseline sharply reduces attention deviation near reusable-block boundaries but leaves interior and task-level residuals, motivating adaptation of the document representation itself. We present \emph{SemPIC}, which trains a LoRA-enabled Writer to compile native per-layer document KVs through behavioral distillation while retaining the pretrained decoder as an unchanged Reader. Adaptation is confined to offline cache construction, preserving the standard KV interface and cache-hit decoding path. We further introduce KV Gradient Checkpointing, which reduces peak training memory without severing gradients through cached KVs. Across three models and four tasks, SemPIC raises mean micro-F1 over KV Packet from 0.53 to 0.60, approaching Full Recompute at 0.62.
\end{abstract}

\section{Introduction}

Retrieval-augmented generation supplies external documents as model context \cite{lewis2020rag}, while tool-using and memory-augmented agents create analogous reusable units. These contexts are long, yet highly repetitive: the query changes from request to request, while many of the underlying semantic units recur. A conventional decoder processes every selected token during prefill before producing the first output token \cite{vaswani2017attention,kwon2023pagedattention}. Prefix caching removes this work only when the reused content appears behind an identical prefix; tree-structured and modular prompt caches broaden sharing, but still rely on constrained layouts \cite{zheng2024sglang,gim2024promptcache}. They cannot fully exploit a document that may be retrieved after different instructions, histories, or neighboring documents.

Position-independent caching (PIC) offers a more general abstraction: prepare the KV cache of each reusable document once, relocate it to its request-time position, and compose it with other cached documents. RoPE re-rotation corrects the positional phase of cached keys \cite{su2024roformer}, but position is only half of the problem. A token's KV state also summarizes the prefix visible when that state was created. A document compiled in isolation never observed the documents that precede it in a later request. Consequently, under a standard causal decoder, naively stitched caches are position aligned but contextually incomplete, and their quality can collapse.

\begin{figure*}[t]
  \centering
  \includegraphics[width=0.9\textwidth]{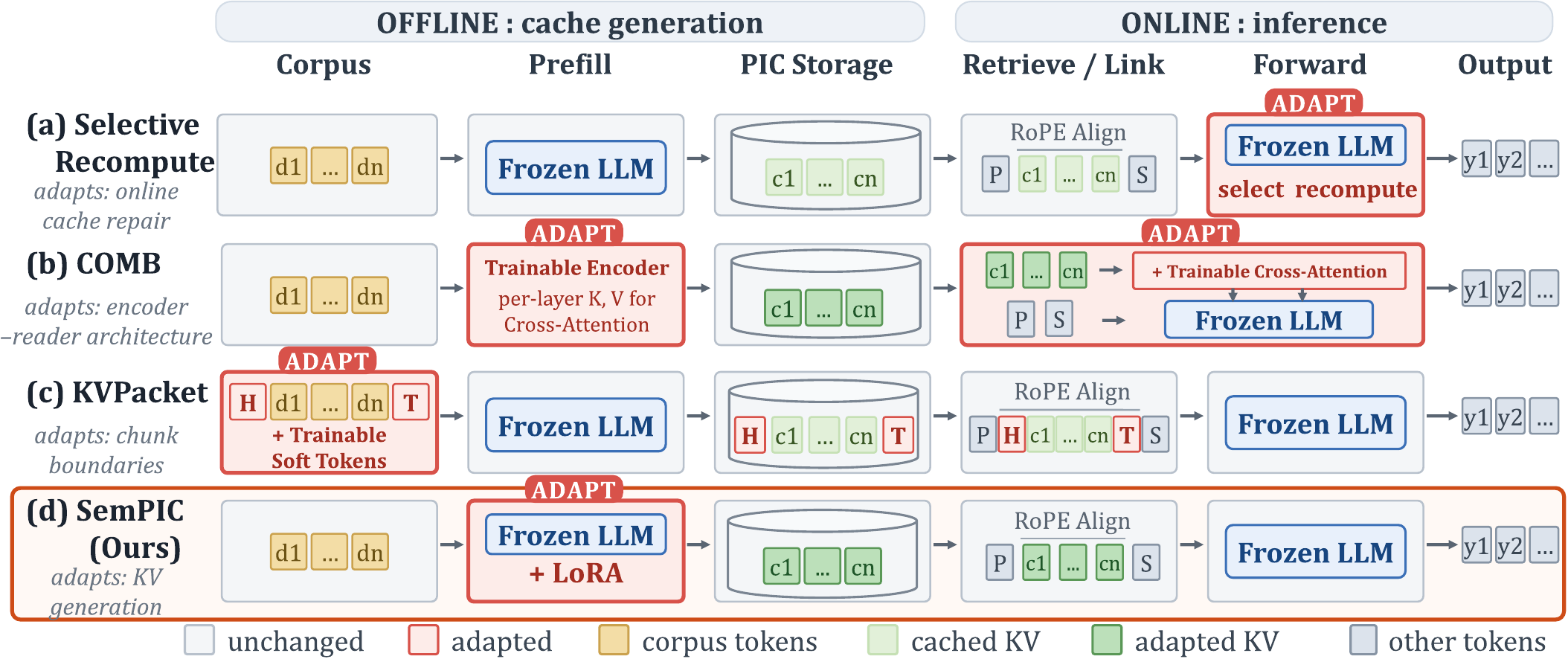}
  \caption{Design space of position-independent cache reuse. Representative designs shown here address PIC's contextual mismatch through online recomputation, offline boundary adaptation, or an auxiliary cache interface. SemPIC broadens the offline trainable locus from boundary embeddings to the native document Writer while retaining standard KVs and an unchanged Reader.}
  \label{fig:overview}
\end{figure*}

\begin{figure*}[t]
  \centering
  \includegraphics[width=0.9\textwidth]{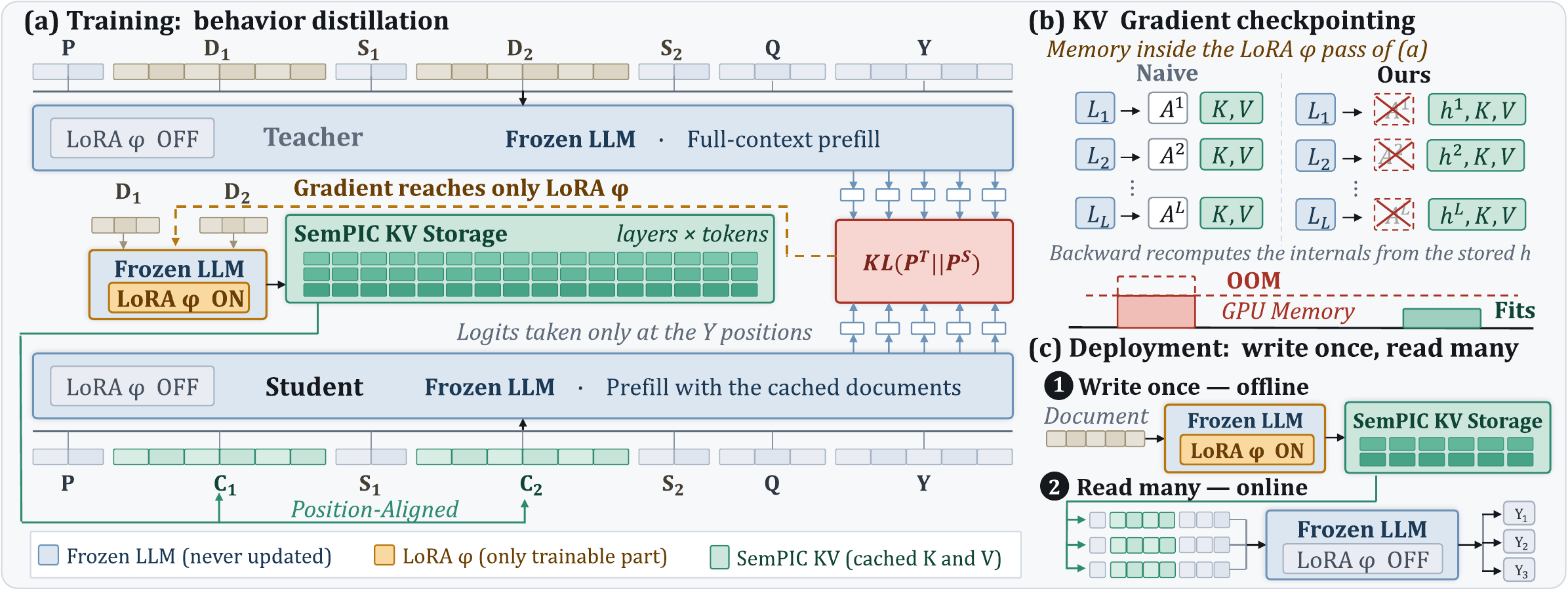}
  \caption{SemPIC trains a document Writer through the model's native KV interface while preserving the pretrained decoder as an unchanged Reader. LoRA is enabled only when the Writer compiles each document; the Reader consumes the differentiable caches with LoRA disabled and matches full-context teacher predictions. Once compiled, the caches enter the standard cache-hit path without document recomputation or auxiliary model execution.}
  \label{fig:training}
\end{figure*}

Existing PIC methods address contextual incompleteness at different loci (Figure~\ref{fig:overview}): CacheBlend and EPIC repair selected document states online \cite{yao2025cacheblend,hu2025epic}, KV Packet learns reusable-block boundary states \cite{chen2026kvpacket}, and COMB adds an auxiliary cache encoder and Reader interface \cite{zhao2026comb}. Yet it remains unclear whether the native document KVs themselves can be learned offline for reuse by an unchanged decoder under unseen document compositions.

To determine where offline adaptation should act, we analyze KV Packet as a boundary-conditioned baseline. It strongly reduces near-boundary attention deviation, but substantial interior deviation and task-level error remain (Section~\ref{sec:motivation}). This boundary--interior asymmetry suggests that boundary states are useful but insufficient, and yields a concrete requirement: adapt the document representation without changing the online cache interface.

SemPIC meets this requirement by broadening adaptation from block boundaries to the document Writer. LoRA \cite{hu2022lora} is active only while the Writer compiles each document, whereas the pretrained Reader consumes the resulting native per-layer KVs with the adapters disabled. Behavioral distillation \cite{chen2026kvpacket} matches the Reader's output distributions to those of Full Recompute, allowing gradients to reshape the Writer through the cache interface while leaving online decoding unchanged.

Making every layer's KVs a differentiable Writer--Reader interface creates a training-memory bottleneck. \textbf{KV Gradient Checkpointing} reconstructs Writer intermediates during backward while preserving the cached KVs and their gradient path, reducing peak training memory without changing cache-hit serving.

Our contributions are:
\begin{itemize}
  \item We identify a consistent \textbf{boundary--interior asymmetry}: boundary conditioning sharply reduces near-boundary attention deviation yet leaves interior and task-level residuals. This observation motivates document-wide cache adaptation.
  \item We propose SemPIC: a trainable native document-to-KV \textbf{Writer} feeds an unchanged pretrained \textbf{Reader} through standard per-layer KVs. Adaptation is offline, and cache-hit decoding remains unmodified.
  \item We introduce \textbf{KV Gradient Checkpointing}, which reconstructs Writer activations while preserving differentiable cache outputs, reducing peak memory enough to train long-context settings that otherwise run out of memory.
  \item Across three models and four tasks, SemPIC improves over KV Packet in 10 of 12 settings, approaching Full Recompute; descriptive diagnostics show lower Full-relative interior attention error in every setting.
\end{itemize}

\section{Background: Position-Independent Caching}

Let a request contain an inline prefix $P$, independently reusable document blocks $D_1,\ldots,D_m$, post-document inline spans $S_1,\ldots,S_m$, request-specific inline query tokens $Q$, and an output prefix $Y_{<t}$. As in Figure~\ref{fig:training}, the canonical order is $P,D_1,S_1,\ldots,D_m,S_m,Q$. The $S_i$ denote dynamic request-time inline spans; unlike $D_i$, they are computed online and are not part of the reusable caches. Full recomputation applies the base model $M_\theta$ to this prompt and defines the teacher distribution
\begin{equation}
\begin{aligned}
p_T^t=p_\theta(y_t\mid {}&P,D_1,S_1,\ldots,D_m,S_m,\\
&Q,Y_{<t}).
\end{aligned}
\end{equation}
A cache compiler $C$ processes each document separately and produces per-layer states $C(D_i)=\{K_i^\ell,V_i^\ell\}_{\ell=1}^{L}$. At request time, a link operator $R$ assigns these states their final logical positions, interleaves them with $S_1,\ldots,S_m$, and applies the causal mask. The resulting student distribution is
\begin{equation}
\begin{aligned}
p_S^t=p_\theta(y_t\mid {}&P,R(C(D_1),S_1,\ldots,\\
&C(D_m),S_m),Q,Y_{<t}).
\end{aligned}
\end{equation}
The PIC objective is not to reconstruct one privileged tensor, but to preserve $p_T$ across future prefixes and document compositions. We call a cache \emph{semantic} when it preserves this downstream predictive behavior, rather than merely approaching a context-specific cache in coordinate space.

For RoPE models, a key created at position $a$ can be moved to $b$ by
\begin{equation}
\widetilde{k}^{\ell}_{b}=R_\ell(b)R_\ell(a)^{-1}k^{\ell}_{a},
\label{eq:rerotate}
\end{equation}
where $R_\ell(\cdot)$ is the layer-specific rotary transform. Values are reused without rotation. Equation~\ref{eq:rerotate} corrects positional phase; it cannot recreate attention to context that was absent during independent compilation.

\section{Related Work}

Selective-recomputation methods repair context-dependent states online. CacheBlend selects tokens by cache deviation, EPIC refreshes leading tokens, and Cache-Craft applies targeted chunk-cache repair \cite{yao2025cacheblend,hu2025epic,agarwal2025cachecraft}. Query-aware selectors refine this choice under different budgets \cite{yang2025cacheclip,wang2026prophetkv,yan2026qcfuse}, but every refreshed token adds work before generation.

Other designs change linking or offline representations. Block-Attention and TurboRAG reuse passage KVs through independent block attention, while KVLink jointly fine-tunes the model and trainable link tokens \cite{ma2025blockattention,lu2025turborag,yang2025kvlink}. KV Packet learns boundary embeddings through self-distillation, COMB adds a document encoder and cross-attention, and recent C$^2$KV learns a compressed cache manifold through a sidecar extractor \cite{chen2026kvpacket,zhao2026comb,du2026c2kv}. SemPIC instead places adaptation in the native document generator: every document token can change its representation, yet the stored object and online Reader remain standard.

This problem is distinct from reducing KV-cache footprint through token eviction or selection, cross-layer merging, quantization, or head-specific caching \cite{zhang2023h2o,li2024snapkv,liu2024minicache,hooper2024kvquant,he2024zipcache,xiao2025duoattention}. SemPIC instead addresses the behavioral mismatch introduced by independent compilation and later composition.

\section{Motivating Analysis: Boundary Conditioning Leaves an Interior Gap}
\label{sec:motivation}

\paragraph{Question.}
KV Packet's boundary-only adaptation suggests an implicit hypothesis: correcting reusable-block interfaces may suffice to restore independently compiled caches. We ask where its gains appear and what mismatch remains.

\paragraph{Diagnostic.}
We compare KV Packet with Full Recompute and No Recompute on Qwen3-4B-Instruct-2507, Qwen3-8B, and Llama-3.1-8B-Instruct across four tasks. Let $D_r(A,B)$ denote the mean absolute deviation between the post-softmax attention probabilities of configurations $A$ and $B$ within normalized block region $r$. We use $r=\mathrm{pre}$ for positions $[0,0.1)$ and $r=\mathrm{int}$ for $[0.1,0.9)$. For $m\in\{\mathrm{Packet},\mathrm{SemPIC}\}$, define its Full-relative error as
\begin{equation}
R_r(m)=\frac{D_r(m,\mathrm{Full})}
{D_r(\mathrm{NR},\mathrm{Full})}.
\label{eq:relative-attention-error}
\end{equation}
Values below one indicate that method $m$ is closer to Full Recompute than No Recompute under this deviation metric. To measure end-task recovery separately, let Vanilla denote direct reuse with No Recompute and define
\begin{equation}
\mathrm{Recovery}(m)=
\frac{\mathrm{F1}_{m}-\mathrm{F1}_{\mathrm{Vanilla}}}
{\mathrm{F1}_{\mathrm{Full}}-\mathrm{F1}_{\mathrm{Vanilla}}}.
\label{eq:f1-recovery}
\end{equation}
Recovery is zero for Vanilla and one for Full Recompute; values above one indicate higher micro-F1 than Full Recompute.

\begin{figure}[t]
  \centering
  \includegraphics[width=0.9\columnwidth]{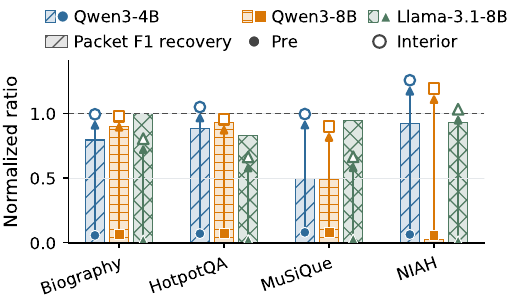}
  \caption{\textbf{KV Packet sharply reduces pre-region error but leaves interior and behavioral residuals.} Hatched bars report KV Packet's F1 recovery toward Full Recompute; arrows connect $R_{\mathrm{pre}}(\mathrm{Packet})$ and $R_{\mathrm{int}}(\mathrm{Packet})$. Marker styles distinguish three models across four tasks. The dashed line at one denotes No Recompute for attention error and Full Recompute for recovery. Attention statistics use 100 samples per model--task configuration.}
  \label{fig:boundary-motivation}
\end{figure}

\paragraph{Observation and design implication.}
Figure~\ref{fig:boundary-motivation} reveals a consistent asymmetry. Across all 12 settings, $R_{\mathrm{pre}}(\mathrm{Packet})$ ranges from 0.016 to 0.081 and remains below one, whereas $R_{\mathrm{int}}(\mathrm{Packet})$ ranges from 0.664 to 1.257 and falls below one in only 8 of 12 settings. KV Packet's mean F1 recovery is 0.763, and its gap to Full Recompute remains positive in 11 settings. Boundary conditioning therefore corrects the near-boundary region without reliably eliminating the interior discrepancy or closing the task-level gap. A reusable-cache compiler should consequently be able to reshape representations throughout the document while preserving an unchanged Reader and the native KV interface. This descriptive analysis motivates the adaptation locus; it does not establish that attention deviation causes the F1 gap.

\section{SemPIC}

The motivating analysis yields a direct design requirement: the offline compiler must support document-wide representation adaptation without changing how caches are consumed online. SemPIC meets this requirement with a trainable Writer and an unchanged Reader connected through native per-layer KVs. We first define this interface and its behavioral supervision, then introduce KV Gradient Checkpointing to make the all-layer gradient path memory efficient.

\subsection{Writer--Reader Learning through the Native KV Interface}

SemPIC separates document representation construction from its subsequent use. A \emph{Writer} processes each reusable document independently, while a \emph{Reader} consumes the compiled documents together with request-specific tokens. Both stages use the same pretrained decoder parameters $\theta$, but only the Writer activates trainable LoRA parameters $\phi$. For document $D_i$, the Writer produces
\begin{equation}
\mathcal{K}_\phi(D_i)=W_{\theta,\phi}(D_i)
=\{K_{i,\phi}^\ell,V_{i,\phi}^\ell\}_{\ell=1}^{L}.
\label{eq:writer}
\end{equation}
The pretrained parameters remain frozen, and $\phi$ denotes the switchable LoRA adaptation activated during Writer compilation. Although only keys and values are exported, the adaptation reshapes the layerwise hidden-state trajectory from which deeper-layer KVs are formed. Equation~\ref{eq:writer} produces the model's native per-layer cache representation, which can be relocated, linked, and consumed by the pretrained decoder without an auxiliary cache format or cross-attention interface.

The Reader is the frozen base model $G_\theta$. After the linker assigns final positions to the compiled caches, it predicts
\begin{equation}
\begin{aligned}
p_S^t=G_\theta\!\bigl(y_t\mid {}&P,
R(\mathcal{K}_\phi(D_1),S_1,\ldots,\\
&\mathcal{K}_\phi(D_m),S_m),Q,Y_{<t}\bigr).
\end{aligned}
\label{eq:reader}
\end{equation}
Crucially, $\phi$ is disabled throughout query prefill, target-token computation, and generation. This asymmetry isolates learning to the document Writer: SemPIC changes the representation placed in the cache without adapting the model that later reads it. The Reader therefore retains the pretrained model's cache layout, positional relocation, and autoregressive decoding semantics.

\paragraph{Behavioral supervision.}

Following KV Packet's self-supervised formulation for cache reuse \cite{chen2026kvpacket}, training applies knowledge distillation \cite{hinton2015distilling} using the base model's full-context behavior as the target. The teacher path supplies detached logits $z_T^{i,r,t}$ for sample $i$, teacher target $r$, and target position $t$. In the student path, the Writer first enables $\phi$ to construct the document caches; the Reader then disables $\phi$ and computes $z_S^{i,r,t}$ from those caches. Disabling the adapter does not detach the KV tensors, so the Reader remains unchanged while its loss still reaches the Writer.

With temperature $\tau$, let $q_T=\operatorname{softmax}(z_T/\tau)$ and $q_S=\operatorname{softmax}(z_S/\tau)$. The implemented objective is
\begin{equation}
\mathcal{L}_{\mathrm{KD}}=\tau^2\sum_i\sum_r\sum_t
D_{\mathrm{KL}}\!\left(q_T^{i,r,t}\,\|\,q_S^{i,r,t}\right).
\label{eq:loss}
\end{equation}
The reported runs use $\tau=1$ and sum the token-level losses. The objective supervises output distributions rather than attention maps, hidden states, or a designated block-interior region. SemPIC's distinction is not the distillation loss itself, but its gradient path and adaptation locus. In particular,
\begin{equation}
\nabla_\phi \mathcal{L}_{\mathrm{KD}}
=\sum_{i,b,\ell}\!\left(
\frac{\partial\mathcal{L}}{\partial K_{i,b}^{\ell}}
\frac{\partial K_{i,b}^{\ell}}{\partial\phi}
+
\frac{\partial\mathcal{L}}{\partial V_{i,b}^{\ell}}
\frac{\partial V_{i,b}^{\ell}}{\partial\phi}
\right),
\label{eq:kv_gradient}
\end{equation}
where $b$ indexes reusable document blocks within sample $i$. Every layer's KVs therefore act simultaneously as the reusable representation consumed by the Reader and the differentiable interface through which the Writer is optimized.

\subsection{KV Gradient Checkpointing for Writer--Reader Training}

KV-mediated Writer--Reader training must retain two classes of state. The Reader requires $\{K^\ell,V^\ell\}_{\ell=1}^{L}$ from the Writer, and these tensors must remain attached to the computation graph so that Equation~\ref{eq:kv_gradient} is valid. A naive autograd-enabled Writer also retains the attention and feed-forward intermediates needed to differentiate each KV tensor with respect to $\phi$. Freezing $\theta$ removes backbone optimizer state, but not these activations. Long documents therefore incur the combined memory cost of the all-layer KV interface and the Writer computation that produced it.

Ordinary activation checkpointing trades retained intermediates for backward recomputation \cite{chen2016sublinear}. SemPIC must make that trade without treating the KVs as disposable layer internals: they cross the Writer--Reader boundary and are the object being learned. We therefore place a checkpoint around each Writer layer with the augmented interface
\begin{equation}
(h^{\ell+1},K^\ell,V^\ell)=C_{\theta,\phi}^{\ell}(h^\ell).
\label{eq:checkpoint}
\end{equation}
The boundary hidden state $h^{\ell+1}$ advances the Writer, while $K^\ell$ and $V^\ell$ remain available to the Reader. Intermediates inside $C_{\theta,\phi}^{\ell}$ are discarded after the forward call and reconstructed only when gradients traverse that layer. The method thus preserves the complete differentiable KV interface but avoids retaining its internal producers across the later Reader computation.

Backward reconstruction may occur after the original Writer forward has completed. Each non-reentrant checkpoint therefore restores the same LoRA-enabled context used to construct the document KVs; otherwise, backward would recompute the Writer layer with LoRA disabled instead of reproducing $C_{\theta,\phi}^{\ell}$. Query-side computation remains LoRA-disabled. KV Gradient Checkpointing thus preserves the Writer function, Reader function, native cache values, and gradient path, changing only the Writer activations retained for backward.

\paragraph{Cache-hit serving.}

Once compiled, SemPIC KVs are ordinary model caches. A cache hit assigns their final logical positions, re-rotates their keys using Equation~\ref{eq:rerotate}, and processes only request-specific inline tokens through the pretrained decoder. Prepared and inline KVs are interleaved under the causal mask before standard autoregressive generation. Query- and layout-independent compilation allows the same cache to be reused across histories, document orders, and retrieved sets, including layouts that place the query at different absolute positions. Semantic adaptation is therefore amortized over cache construction and introduces neither a document-side forward pass nor an auxiliary model on the request critical path.

\paragraph{Composition with boundary adaptation.}

To test whether document-wide and boundary adaptation are complementary, \emph{Joint} co-trains shared Header and Trailer embeddings with the LoRA-enabled Writer under the same behavioral-distillation loss while the pretrained model remains frozen. The Writer adapts the document representation, while the boundary states provide a learned block interface. Once compiled, Joint follows SemPIC's recomputation-free cache-hit path.

\section{Experiments}

We organize the evaluation around six questions: Does semantic compilation recover task quality? What quality--efficiency tradeoff does it offer? Does it reduce the interior attention discrepancy that motivated the method? Does the learned Writer transfer across domains? Which block-local artifacts remain after adaptation? What memory--time tradeoff does KV Gradient Checkpointing introduce?

\subsection{Setup}

We evaluate Llama-3.1-8B-Instruct \cite{grattafiori2024llama}, Qwen3-4B-Instruct-2507, and Qwen3-8B \cite{yang2025qwen3} on Synthetic Biographies \cite{karev2025biographies}, HotpotQA \cite{yang2018hotpotqa}, MuSiQue \cite{trivedi2022musique}, and a Needle-in-a-haystack (NIAH) task. We choose these tasks to span complementary cache-composition regimes: Biography tests factual retrieval from short reusable records, HotpotQA and MuSiQue require multi-hop QA over evidence distributed across multiple documents, and NIAH isolates long-context retrieval of sparse answer-bearing evidence. The latter is a controlled retrieval diagnostic of the type included in long-context suites such as RULER \cite{hsieh2024ruler}. We train SemPIC on training partitions and evaluate it on held-out test partitions. Each model--task--method cell contains 100 examples selected with a fixed dataset seed, greedy decoding, and bfloat16 execution; task-specific generation limits are reported in the supplementary material. SemPIC adapters are trained per model and domain for five epochs with LoRA rank 8, scale 16, and a learning rate of $5\times10^{-4}$ with linear decay; complete settings are in the supplementary material.

Full Recompute processes the complete prompt and serves as the quality reference. No Recompute directly links independently prepared KVs after positional realignment, while No Cache omits the retrieved documents as a model-only control. KV Packet learns Header and Trailer tokens around document tokens\cite{chen2026kvpacket}; Joint is the co-trained composition of these boundary embeddings with the SemPIC LoRA described above. We report corpus-level token micro-F1.

\subsection{Semantic Compilation Recovers PIC Quality}

\begin{table}[t]
\centering
\small
\begin{tabular*}{0.9\columnwidth}{@{\extracolsep{\fill}}lccc>{\columncolor{black!6}}c>{\columncolor{black!6}}c@{}}
\toprule
Task & Full & NR & Packet & \textbf{SemPIC} & \textbf{Joint} \\
\midrule
\multicolumn{6}{l}{\textbf{Llama-3.1-8B}} \\
Biography & 0.96 & 0.06 & 0.96 & \textbf{0.99} & \textbf{0.99} \\
HotpotQA & 0.48 & 0.24 & 0.44 & \textbf{0.45} & 0.41 \\
MuSiQue & 0.39 & 0.01 & 0.37 & 0.28 & \textbf{0.38} \\
NIAH & 0.83 & 0.39 & 0.80 & \textbf{0.82} & \textbf{0.82} \\
\textbf{Average} & 0.67 & 0.18 & 0.64 & 0.64 & \textbf{0.65} \\
\midrule
\multicolumn{6}{l}{\textbf{Qwen3-4B}} \\
Biography & 0.97 & 0.19 & 0.81 & \textbf{0.98} & 0.96 \\
HotpotQA & 0.31 & 0.14 & 0.29 & 0.29 & \textbf{0.30} \\
MuSiQue & 0.41 & 0.07 & 0.24 & 0.31 & \textbf{0.35} \\
NIAH & 0.67 & 0.17 & 0.63 & \textbf{0.76} & 0.74 \\
\textbf{Average} & 0.59 & 0.14 & 0.49 & \textbf{0.59} & \textbf{0.59} \\
\midrule
\multicolumn{6}{l}{\textbf{Qwen3-8B}} \\
Biography & 0.96 & 0.24 & 0.89 & \textbf{0.98} & 0.97 \\
HotpotQA & 0.28 & 0.13 & 0.27 & 0.28 & \textbf{0.29} \\
MuSiQue & 0.45 & 0.08 & 0.26 & 0.34 & \textbf{0.38} \\
NIAH & 0.72 & 0.37 & 0.38 & \textbf{0.70} & \textbf{0.70} \\
\textbf{Average} & 0.60 & 0.21 & 0.45 & 0.58 & \textbf{0.59} \\
\midrule
\shortstack[l]{\textbf{Overall}\\\textbf{Average}} & 0.62 & 0.17 & 0.53 & 0.60 & \textbf{0.61} \\
\bottomrule
\end{tabular*}
\caption{Task-level corpus micro-F1. NR and Packet denote No Recompute and KV Packet. Gray columns mark SemPIC and Joint; bold marks the strongest learned-cache result in each row. Average and Overall Average are arithmetic means over four tasks and all 12 task rows, respectively.}
\label{tab:main}
\end{table}

Table~\ref{tab:main} shows higher micro-F1 for SemPIC than KV Packet in 10 of 12 settings, lifting the overall mean from 0.53 to 0.60, close to Full Recompute at 0.62. SemPIC matches or exceeds KV Packet across all eight Qwen3 settings, improving the 4B and 8B model averages by 0.10 and 0.13. The exception is Llama MuSiQue (0.28 versus 0.37), where Joint reaches 0.38; Joint's overall mean of 0.61 indicates that document-wide and boundary adaptation can be complementary. Because relocation, linking, and online decoding are shared, the central experimental contrast is the adaptation locus: document-wide for SemPIC and boundary-only for KV Packet.

\begin{figure*}[t]
  \centering
  \includegraphics[width=0.9\textwidth]{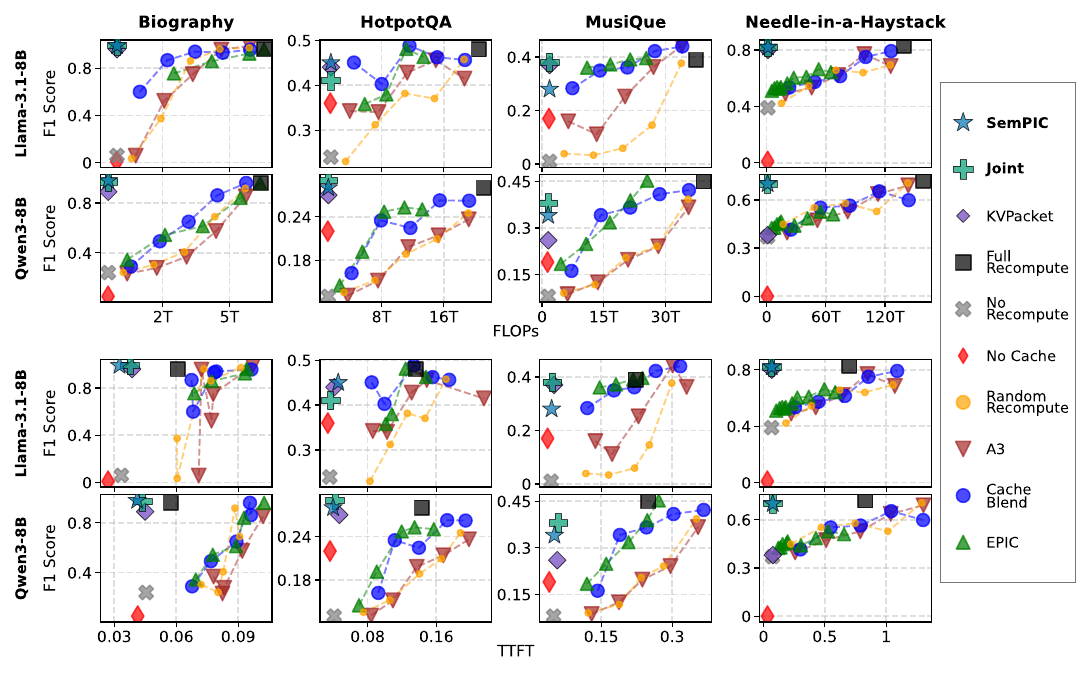}
  \caption{Quality--efficiency tradeoffs on Llama-3.1-8B and Qwen3-8B across four tasks. The upper two rows plot corpus micro-F1 against mean online prefill FLOPs; the lower two rows plot F1 against cache-hit time to first token (TTFT). Curves sweep selective-recomputation budgets, while isolated markers denote learned recomputation-free methods or references. Qwen3-4B and an enlarged three-model view are reported in the supplementary material.}
  \label{fig:tradeoff}
\end{figure*}

\subsection{Quality--Efficiency Tradeoff}

Offline learned-cache methods and online selective-recomputation methods occupy complementary operating regimes (Figure~\ref{fig:tradeoff}). Learned-cache adaptation is paid offline, whereas selective methods return document computation to the request path. No method dominates across all tasks; the preferred regime depends on whether a deployment permits online recomputation.

Following KV Packet's comparison grid \cite{chen2026kvpacket}, CacheBlend \cite{yao2025cacheblend}, A$^3$ \cite{zhou2025a3}, and Random Recompute sweep 10\%, 30\%, 50\%, 70\%, and 90\% of document tokens. EPIC \cite{hu2025epic} recomputes 10, 30, 50, 70, and 90 tokens, with budgets from 100 to 500 tokens for NIAH. Figure~\ref{fig:tradeoff} orders each sweep by increasing recomputation budget.

\begin{figure}[!t]
  \centering
  \includegraphics[width=0.9\columnwidth]
    {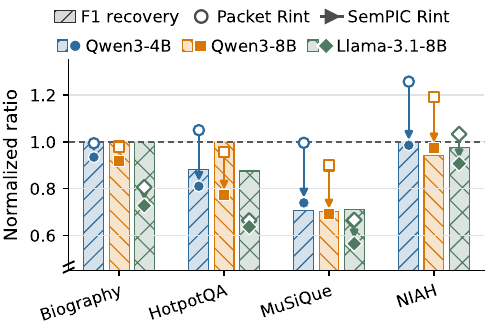}
  \caption{\textbf{SemPIC lowers Full-relative interior attention error in all 12 settings.}
  Hatched bars show F1 recovery from No Recompute toward Full Recompute,
  capped at one for visualization; text reports unclipped values.
  Arrows connect $R_{\mathrm{int}}(\mathrm{Packet})$ and
  $R_{\mathrm{int}}(\mathrm{SemPIC})$, where lower is better.
  Results cover three models and four tasks, with 100 samples per configuration.}
  \label{fig:interior-validation}
\end{figure}

\subsection{Interior Attention Analysis}

Across all 12 settings, SemPIC reduces $R_{\mathrm{int}}(\mathrm{SemPIC})$ below both the No Recompute reference and $R_{\mathrm{int}}(\mathrm{Packet})$, ranging from 0.564 to 0.985 versus 0.664 to 1.257 for KV Packet (Figure~\ref{fig:interior-validation}). Its unclipped recovery toward Full Recompute ranges from 0.703 to 1.180, with a mean of 0.921. The lower interior error and stronger task-level results relative to KV Packet move in the same favorable direction, but the evidence remains descriptive and does not show that the attention change causes the F1 gain.

\subsection{Cross-Domain Transfer}

We next test whether a Writer transfers across document domains.

On Qwen3-8B, SemPIC improves KV Packet's four-target average for every training source. Joint attains the highest average for four of the five sources, counting ties on MuSiQue and Mixture. Mixture training yields the highest average for each method: 0.51 for KV Packet and 0.57 for both SemPIC and Joint (Table~\ref{tab:transfer}). Across all three models, SemPIC improves the corresponding KV Packet average under 14 of 15 training sources; the complete Llama-3.1-8B and Qwen3-4B matrices are reported in the supplementary material.

\begin{table}[h]
\centering
\small
\begin{tabular*}{0.9\columnwidth}{@{\extracolsep{\fill}}lccccc@{}}
\toprule
\textbf{Method} & \shortstack{\textbf{Bio-}\\\textbf{graphy}} & \shortstack{\textbf{Hotpot}\\\textbf{QA}} & \shortstack{\textbf{Mu}\\\textbf{SiQue}} & \textbf{NIAH} & \shortstack{\textbf{Aver-}\\\textbf{age}} \\
\midrule
\multicolumn{6}{l}{\textbf{Train: Biography}} \\
Packet & 0.89 & 0.17 & 0.07 & 0.44 & 0.39 \\
\rowcolor{black!6}
\textbf{SemPIC} & 0.98 & 0.16 & 0.12 & 0.32 & 0.40 \\
\rowcolor{black!6}
\textbf{Joint} & 0.97 & 0.22 & 0.16 & 0.39 & \textbf{0.44} \\
\midrule
\multicolumn{6}{l}{\textbf{Train: HotpotQA}} \\
Packet & 0.73 & 0.27 & 0.29 & 0.41 & 0.43 \\
\rowcolor{black!6}
\textbf{SemPIC} & 0.86 & 0.28 & 0.38 & 0.46 & \textbf{0.50} \\
\rowcolor{black!6}
\textbf{Joint} & 0.85 & 0.29 & 0.37 & 0.37 & 0.47 \\
\midrule
\multicolumn{6}{l}{\textbf{Train: MuSiQue}} \\
Packet & 0.75 & 0.24 & 0.26 & 0.47 & 0.43 \\
\rowcolor{black!6}
\textbf{SemPIC} & 0.76 & 0.31 & 0.34 & 0.45 & \textbf{0.47} \\
\rowcolor{black!6}
\textbf{Joint} & 0.83 & 0.27 & 0.38 & 0.40 & \textbf{0.47} \\
\midrule
\multicolumn{6}{l}{\textbf{Train: NIAH}} \\
Packet & 0.53 & 0.18 & 0.19 & 0.38 & 0.32 \\
\rowcolor{black!6}
\textbf{SemPIC} & 0.25 & 0.21 & 0.22 & 0.70 & 0.35 \\
\rowcolor{black!6}
\textbf{Joint} & 0.36 & 0.21 & 0.18 & 0.70 & \textbf{0.36} \\
\midrule
\multicolumn{6}{l}{\textbf{Train: Mixture}} \\
Packet & 0.94 & 0.23 & 0.31 & 0.56 & 0.51 \\
\rowcolor{black!6}
\textbf{SemPIC} & 0.96 & 0.26 & 0.36 & 0.70 & \textbf{0.57} \\
\rowcolor{black!6}
\textbf{Joint} & 0.97 & 0.25 & 0.36 & 0.71 & \textbf{0.57} \\
\bottomrule
\end{tabular*}
\caption{Cross-domain corpus micro-F1 on Qwen3-8B. Packet denotes KV Packet. Each group identifies the training source and columns identify the evaluation target; Average is their arithmetic mean. Gray rows mark SemPIC and Joint; bold numeric entries mark the highest average within each training-source group.}
\label{tab:transfer}
\end{table}

\subsection{Residual Block-Local Behavior and Scope}

Attention sinks often concentrate on sequence-initial tokens \cite{xiao2024streamingllm,gu2025attentionsink}. Independent compilation places every document's first token at block-local offset one, so we test for a corresponding concentration after linking. For block token offset $j$, let $\operatorname{density}_{\mathrm{SemPIC}}(j)$ denote its attention density; we normalize it by the width-weighted mean density over the block interior $[0.1,0.9)$:
\begin{equation}
T_j=\frac{\operatorname{density}_{\mathrm{SemPIC}}(j)}
  {\operatorname{density}_{\mathrm{SemPIC}}([0.1,0.9))}.
\label{eq:block-token-ratio}
\end{equation}

\begin{figure}[!t]
  \centering
  \includegraphics[width=0.9\columnwidth]{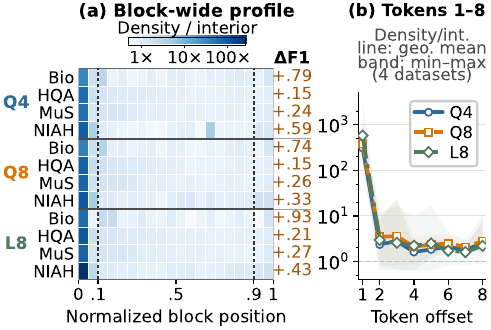}
  \captionof{figure}{\textbf{SemPIC's behavioral improvements coexist with a block-local first-token peak.} (a) Normalized block-position attention density with $\Delta\mathrm{F1}=\mathrm{F1}_{\mathrm{SemPIC}}-\mathrm{F1}_{\mathrm{NR}}$ labels. (b) $T_j$ at offsets 1--8; lines and bands give geometric means and task ranges. Q4/Q8/L8 denote Qwen3-4B/Qwen3-8B/Llama-3.1-8B; $n=100$ per configuration.}
  \label{fig:block-token-peak}
\end{figure}

Across all 12 settings, $T_1$ is the largest of the first eight ratios, ranging from 132.5 to 3248.3 versus 0.61 to 23.49 for offsets 2--8 (Figure~\ref{fig:block-token-peak}). The peak is localized at block offset one rather than extending across neighboring offsets. This descriptive diagnostic does not separate block position from repeated block-start content, but it shows that better task behavior can coexist with a strong block-local concentration.

\subsection{Training Memory and Time}

Table~\ref{tab:memory} compares KV Gradient Checkpointing with the naive training path in peak device memory and wall-clock time.

\begin{center}
\centering
\small
\begin{tabular*}{0.9\columnwidth}{@{\extracolsep{\fill}}lrrrr@{}}
\toprule
& \multicolumn{2}{c}{Naive} & \multicolumn{2}{c}{Ours} \\
\cmidrule(lr){2-3}\cmidrule(lr){4-5}
Dataset & Mem. & Time & Mem. & Time \\
\midrule
Biography & 23.8 & 0:34 & 18.0 & 1:55 \\
Biography (FBS 8) & 75.8 & 0:23 & 33.2 & 0:25 \\
HotpotQA & \multicolumn{2}{c}{OOM} & 26.4 & 5:40 \\
MuSiQue & \multicolumn{2}{c}{OOM} & 33.1 & 6:46 \\
NIAH & \multicolumn{2}{c}{OOM} & 43.4 & 13:26 \\
\bottomrule
\end{tabular*}
\captionof{table}{Qwen3-8B SemPIC peak training memory (GiB) and wall-clock time (h:mm) on one A100 80GB GPU. The document-forward batch size (FBS) is 1 unless marked FBS 8; Ours denotes KV Gradient Checkpointing.}
\label{tab:memory}
\end{center}

KV Gradient Checkpointing lowers Qwen3-8B peak memory from 23.8 to 18.0 GiB at FBS 1 and from 75.8 to 33.2 GiB at FBS 8. The corresponding Biography times increase from 34 to 115 minutes at FBS 1 and from 23 to 25 minutes at FBS 8, showing substantial checkpointing overhead at the smaller microbatch but little at FBS 8. On HotpotQA, MuSiQue, and NIAH, the checkpointed path completes where the naive implementation runs out of memory. These measurements characterize a memory--time tradeoff, not a general throughput claim; cache-hit serving is unchanged.

\section{Conclusion}

Position-independent reuse asks a document cache to function in contexts absent during construction. SemPIC addresses this mismatch by training a Writer to produce native KVs for an unchanged Reader. Across the evaluated settings, SemPIC raises overall mean micro-F1 over KV Packet from 0.53 to 0.60 and reduces Full-relative interior attention error under our metric. Joint suggests that document-wide and boundary adaptation are complementary, while KV Gradient Checkpointing makes long-document Writer training feasible.

\bibliography{sempic}

\clearpage
\appendix
\suppressfloats[t]
\section*{Supplementary Material}
This supplement provides implementation, measurement, and reproducibility details for the main paper. It does not add claims beyond the reported experimental scope.

\section{SemPIC Training Details}

\subsection{Document and Student Paths}

For each training sample, SemPIC identifies prompt spans designated as reusable context blocks. The frozen token embedding maps every block to input embeddings, and the document adapter encodes the blocks independently. Independence is maintained under batching: padding is masked, and a block cannot attend to another block in the same batch.

Prepared KVs are assigned to each associated teacher target and linked with the inline query tokens and teacher-forced output prefix. Student logits are collected only at target-token positions. The base-model parameters remain frozen, the adapter is disabled during student computation, and cached teacher logits are detached. No teacher forward is executed inside the measured student-training process.

\subsection{Dataset Splits and Sample Construction}

Examples are split before prompt construction. Biography and NIAH use disjoint project-level training and test partitions; HotpotQA and MuSiQue use the official training and validation splits. Distillation contains 1,024 Biography records---256 each with 2, 3, 5, and 7 documents---and 2,048 records per remaining task; evaluation uses 100 held-out zero-shot examples per task.

\subsection{Loss and Optimization}

The teacher cache contains one or more generated sequences per sample and a floating-point logit tensor over the complete vocabulary for every generated token. The temperature-scaled forward KL objective defined in the main paper is implemented by summing over the complete vocabulary, target-token, teacher-target, and sample dimensions. The evaluated runs use $\tau=1$.

Table~\ref{tab:train-config} records the common adapter settings. Adapters are trained separately by model and domain.

\begin{table}[t]
\centering
\begin{tabular}{ll}
\toprule
Setting & Value \\
\midrule
Adapter & LoRA on q/k/v/o projections \\
Rank / scale / dropout & 8 / 16 / 0 \\
Epochs & 5 \\
Learning rate & $5\times10^{-4}$ \\
Schedule & Linear decay to zero \\
Weight decay & 0 \\
Precision & bfloat16 \\
Loss temperature & 1 \\
Evaluation-data seed & 42 \\
\bottomrule
\end{tabular}
\caption{Common SemPIC training and evaluation settings for the evaluated campaign.}
\label{tab:train-config}
\end{table}

\section{KV Gradient Checkpointing}

\subsection{Layer-Wise Contract}

The checkpointed Writer uses the layer-wise hidden-state and KV interface defined in the main paper. Each non-reentrant checkpoint receives the same causal mask, position identifiers, cache positions, rotary embeddings, and model-specific sliding-window mask as the direct path. Variable-length blocks are padded during batched construction and trimmed to their valid lengths afterward. A context factory re-enters the LoRA-enabled Writer state during both original execution and backward recomputation.

The implementation supports the Llama and Qwen3 decoder families used in the paper. It requires document input embeddings and an autograd-enabled document-KV forward. The current training path does not combine checkpointed document construction with KV compression or index-based KV selection.

\subsection{Numerical Validation}

Direct and checkpointed paths are tested for per-layer key/value equality, position handling, and input gradients on variable-length Llama and Qwen3 examples. Student-prefill tests verify that LoRA gradients survive checkpointing. Joint-method tests additionally compare losses and gradients when packet parameters and LoRA are trained together. These checks validate the forward and gradient contract; they do not substitute for the memory measurements below.

\subsection{Memory Measurement Protocol}

The main comparison uses Qwen3-8B LoRA-only KL training on one NVIDIA A100 80GB PCIe GPU per process. Each dataset is measured once with checkpointing enabled and once with it disabled. The forward microbatch contains one sample, while the optimizer batch contains 64 samples accumulated across microbatches. The model executes in bfloat16 with rank-8 LoRA on q/k/v/o projections.

Device memory is sampled with \texttt{nvidia-smi} every two seconds. Each GPU is empty before launch and hosts only the measured process. A successful probe enters the real forward/backward training path; selected probes complete at least one optimizer update. A condition is recorded as OOM when the normal configuration cannot complete its forward/backward path. The values include model weights, LoRA state, runtime buffers, and resident teacher-generation data, so they are system-level device observations rather than isolated activation allocations.

The resulting microbatch-1 measurements are summarized in the main paper. A separate Biography probe uses document-forward microbatch size 8. Its direct path operates near device capacity and emits allocation-pressure warnings, so this condition is kept separate from the microbatch-1 cross-dataset comparison.

These are short operational probes rather than repeated full training runs. The two-second sampling period may miss short-lived allocator peaks, and no variance estimate is available. They support the fit-versus-OOM and approximate memory-effect claims, not a statistical characterization of GPU memory.

\section{Position-Independent Linking}

\subsection{Logical Layout and Re-Rotation}

Prepared blocks can be stored and batched independently of their later prompt order. At link time, the runtime builds a token-level map from physical cache slots to logical prompt positions, applies the key re-rotation defined in the main paper, and reuses values directly. Inline positions are processed layer by layer and their new KVs are interleaved with the prepared document states. The same caches therefore support different queries, histories, document orders, and retrieved sets, with logical positions assigned for each layout.

A frontier causal mask prevents each inline token from observing logical successors, even when physical slots for later blocks have already been assembled. The terminal inline hidden state is projected through the language-model head to obtain the first-token distribution. The completed cache is then used by ordinary autoregressive decoding.

\subsection{TTFT Boundary}

The paper reports cache-hit online TTFT. The timer starts after prepared document KVs and method-specific wrappers are available, immediately before prompt-layout construction. It ends after the first output token is produced, with device synchronization at both endpoints. It includes:
\begin{itemize}
  \item logical layout and causal-frontier construction;
  \item document-key re-rotation;
  \item inline-token prefill and cache assembly;
  \item generation setup and first-token generation.
\end{itemize}
It excludes model loading, data preprocessing, cache lookup and transfer I/O, offline document-KV construction, and cache-miss ingestion. This boundary models repeated cache-hit service rather than first use of a new document.

\section{Evaluation Details}

\subsection{Quality Metric}

After task-specific answer normalization, each prediction and reference is split on whitespace and treated as a token multiset. True positives, false positives, and false negatives are pooled over the complete 100-example set, with $P=TP/(TP+FP)$, $R=TP/(TP+FN)$, and $F_1=2PR/(P+R)$. We choose corpus micro-F1 because these retrieval and open-form QA tasks can return multi-token answers: token overlap credits partially recovered answer content, while precision penalizes extra non-reference tokens, rather than assigning every non-exact generation zero credit. The metric remains lexical, whitespace-sensitive, and order-insensitive.

All methods use greedy decoding. Generation caps are 32/512/128 tokens for Biography/HotpotQA/NIAH; MuSiQue uses 256 for Llama-3.1-8B and 512 for Qwen3.

\subsection{Cross-Domain Transfer}

Tables~\ref{tab:transfer-llama} and~\ref{tab:transfer-qwen4} complete the main-paper transfer analysis under the same training-source and evaluation-target organization. SemPIC improves the corresponding KV Packet four-target average under all five Llama-3.1-8B training sources and four of five Qwen3-4B sources; the exception is Qwen3-4B trained on MuSiQue.

\begin{table}[t]
\centering
\small
\begin{tabular*}{0.9\columnwidth}{@{\extracolsep{\fill}}lccccc@{}}
\toprule
\textbf{Method} & \shortstack{\textbf{Bio-}\\\textbf{graphy}} & \shortstack{\textbf{Hotpot}\\\textbf{QA}} & \shortstack{\textbf{Mu}\\\textbf{SiQue}} & \textbf{NIAH} & \shortstack{\textbf{Aver-}\\\textbf{age}} \\
\midrule
\multicolumn{6}{l}{\textbf{Train: Biography}} \\
Packet & 0.96 & 0.18 & 0.16 & 0.44 & 0.44 \\
\textbf{SemPIC} & 0.99 & 0.32 & 0.10 & 0.40 & \textbf{0.45} \\
\textbf{Joint} & 0.99 & 0.26 & 0.08 & 0.40 & 0.43 \\
\midrule
\multicolumn{6}{l}{\textbf{Train: HotpotQA}} \\
Packet & 0.88 & 0.44 & 0.39 & 0.36 & 0.52 \\
\textbf{SemPIC} & 0.95 & 0.45 & 0.33 & 0.56 & 0.57 \\
\textbf{Joint} & 0.96 & 0.41 & 0.38 & 0.56 & \textbf{0.58} \\
\midrule
\multicolumn{6}{l}{\textbf{Train: MuSiQue}} \\
Packet & 0.87 & 0.36 & 0.37 & 0.34 & 0.49 \\
\textbf{SemPIC} & 0.95 & 0.44 & 0.28 & 0.54 & 0.55 \\
\textbf{Joint} & 0.92 & 0.37 & 0.38 & 0.64 & \textbf{0.58} \\
\midrule
\multicolumn{6}{l}{\textbf{Train: NIAH}} \\
Packet & 0.54 & 0.28 & 0.07 & 0.80 & 0.42 \\
\textbf{SemPIC} & 0.53 & 0.36 & 0.22 & 0.82 & 0.48 \\
\textbf{Joint} & 0.85 & 0.34 & 0.06 & 0.82 & \textbf{0.52} \\
\midrule
\multicolumn{6}{l}{\textbf{Train: Mixture}} \\
Packet & 0.95 & 0.42 & 0.43 & 0.71 & 0.63 \\
\textbf{SemPIC} & 0.99 & 0.42 & 0.38 & 0.77 & 0.64 \\
\textbf{Joint} & 0.99 & 0.49 & 0.39 & 0.82 & \textbf{0.67} \\
\bottomrule
\end{tabular*}
\caption{Cross-domain corpus micro-F1 on Llama-3.1-8B. Packet denotes KV Packet. Each group identifies the training source and columns identify the evaluation target; Average is their arithmetic mean. Bold numeric entries mark the best displayed average for each source.}
\label{tab:transfer-llama}
\end{table}

\begin{table}[t]
\centering
\small
\begin{tabular*}{0.9\columnwidth}{@{\extracolsep{\fill}}lccccc@{}}
\toprule
\textbf{Method} & \shortstack{\textbf{Bio-}\\\textbf{graphy}} & \shortstack{\textbf{Hotpot}\\\textbf{QA}} & \shortstack{\textbf{Mu}\\\textbf{SiQue}} & \textbf{NIAH} & \shortstack{\textbf{Aver-}\\\textbf{age}} \\
\midrule
\multicolumn{6}{l}{\textbf{Train: Biography}} \\
Packet & 0.81 & 0.18 & 0.07 & 0.31 & 0.34 \\
\textbf{SemPIC} & 0.98 & 0.20 & 0.13 & 0.44 & \textbf{0.44} \\
\textbf{Joint} & 0.96 & 0.17 & 0.13 & 0.21 & 0.37 \\
\midrule
\multicolumn{6}{l}{\textbf{Train: HotpotQA}} \\
Packet & 0.81 & 0.29 & 0.28 & 0.50 & 0.47 \\
\textbf{SemPIC} & 0.89 & 0.29 & 0.30 & 0.55 & \textbf{0.51} \\
\textbf{Joint} & 0.78 & 0.30 & 0.37 & 0.49 & 0.49 \\
\midrule
\multicolumn{6}{l}{\textbf{Train: MuSiQue}} \\
Packet & 0.67 & 0.26 & 0.24 & 0.48 & 0.41 \\
\textbf{SemPIC} & 0.49 & 0.18 & 0.31 & 0.49 & 0.37 \\
\textbf{Joint} & 0.67 & 0.26 & 0.35 & 0.49 & \textbf{0.44} \\
\midrule
\multicolumn{6}{l}{\textbf{Train: NIAH}} \\
Packet & 0.56 & 0.14 & 0.12 & 0.63 & 0.36 \\
\textbf{SemPIC} & 0.27 & 0.26 & 0.18 & 0.76 & 0.37 \\
\textbf{Joint} & 0.47 & 0.25 & 0.19 & 0.74 & \textbf{0.41} \\
\midrule
\multicolumn{6}{l}{\textbf{Train: Mixture}} \\
Packet & 0.93 & 0.26 & 0.24 & 0.48 & 0.48 \\
\textbf{SemPIC} & 0.94 & 0.28 & 0.42 & 0.67 & \textbf{0.58} \\
\textbf{Joint} & 0.96 & 0.27 & 0.34 & 0.75 & \textbf{0.58} \\
\bottomrule
\end{tabular*}
\caption{Cross-domain corpus micro-F1 on Qwen3-4B-Instruct-2507. Packet denotes KV Packet. Each group identifies the training source and columns identify the evaluation target; Average is their arithmetic mean. Bold numeric entries mark the best displayed average for each source.}
\label{tab:transfer-qwen4}
\end{table}

\subsection{Efficiency Metrics}

Mean online prefill FLOPs are architecture-aware analytical estimates averaged over the 100 evaluation requests. For cache-reuse methods, accounting begins after reusable KVs and learned wrappers are prepared; Full Recompute instead counts complete-prompt prefill. The estimate includes key re-rotation, transformer arithmetic for inline and recomputed tokens, implemented method-specific scoring or fusion, and the final normalization and output projection that produce first-token logits. It excludes offline cache construction and training, model loading, data preprocessing, cache lookup and transfer, and autoregressive computation after the first-token logits. It also omits system effects not represented by the analytical counters, including data movement, allocation, layout construction, kernel launch, and synchronization overhead.

TTFT is the mean synchronized wall-clock time under the boundary defined above; for cache-reuse methods, this is cache-hit latency. We report both metrics because FLOPs isolates algorithmic online arithmetic independently of a particular GPU implementation, whereas TTFT captures realized serving effects such as parallelism, memory traffic, and kernel efficiency. Conversely, the FLOP axis helps distinguish reduced online computation from implementation-specific latency differences.

\subsection{Computing Environment}

All quality and TTFT comparisons use the same server and software environment. The server contains eight NVIDIA A100 80GB PCIe GPUs, two Intel Xeon Gold 6530 CPUs (64 physical cores and 128 hardware threads), 1~TiB of host memory, and Ubuntu 24.04.3 LTS; each process uses one GPU. The environment uses Python 3.12.3, CUDA 13.0, PyTorch 2.13.0+cu130, Transformers 5.2.0, PEFT 0.19.1, Accelerate 1.14.0, and Datasets 5.0.0, with NVIDIA driver 610.43.02. All methods within a TTFT panel are therefore compared under matched hardware and software conditions.

\subsection{Complete High-Resolution Quality--Efficiency Overview}

\begin{figure*}[p]
  \centering
  \includegraphics[width=0.885\textwidth]{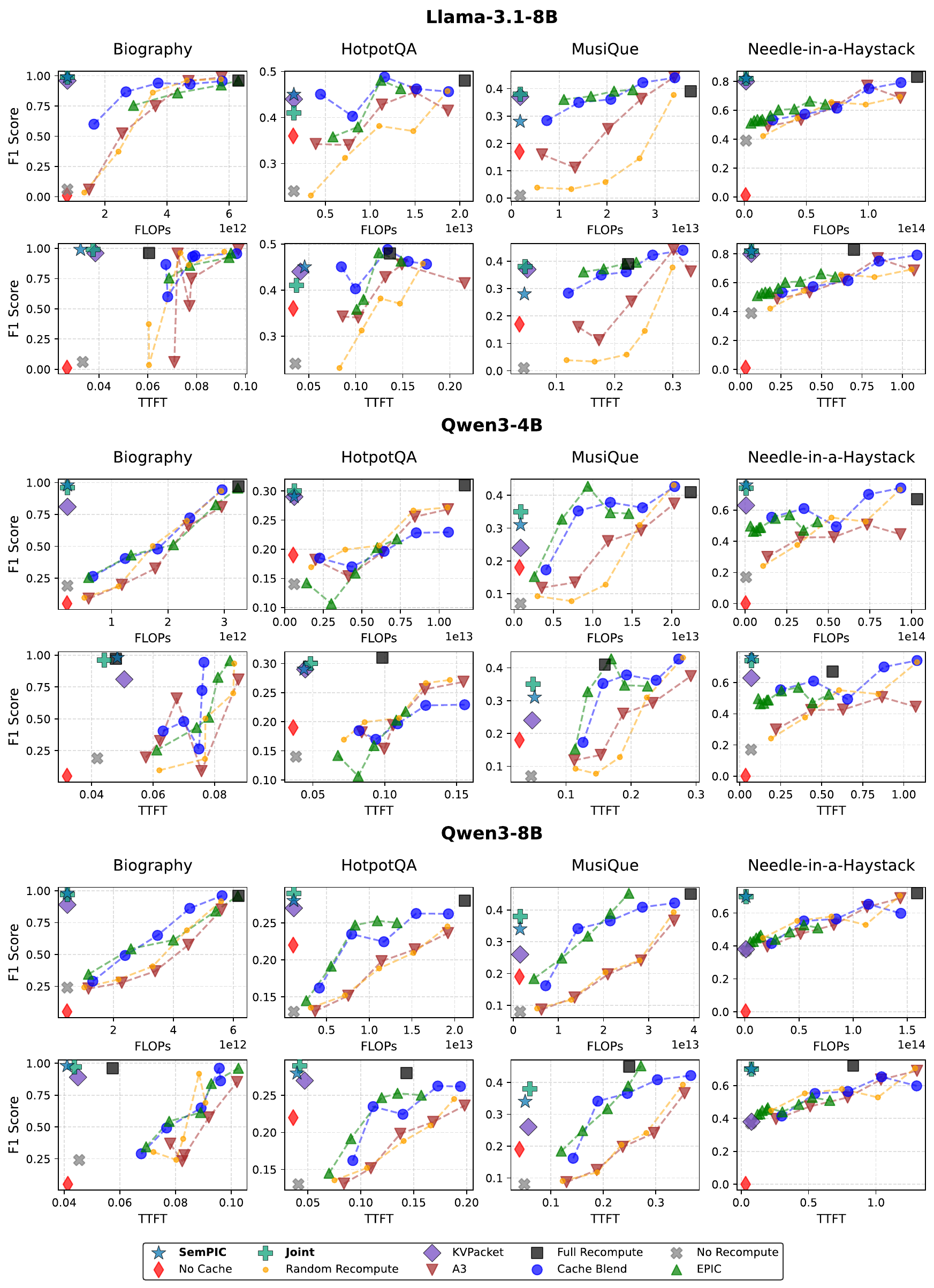}
  \caption{High-resolution overview of quality--efficiency tradeoffs on three models and four tasks, reporting corpus micro-F1 against online prefill FLOPs (upper rows) and cache-hit TTFT (lower rows).}
  \label{fig:tradeoff-all}
\end{figure*}

Figure~\ref{fig:tradeoff-all} provides a single high-resolution overview of all three models under the same four-task protocol. Within each model block, the upper row uses mean online prefill FLOPs and the lower row uses cache-hit TTFT, with selective-recomputation sweeps and recomputation-free learned-cache points shown together. TTFT excludes offline cache construction. This view is descriptive: it exposes the available operating points under a common metric boundary without treating connected budget sweeps as continuous curves or repeated trials.

\subsection{Interpretation Boundaries}

The reported tradeoff curves do not establish performance at unevaluated recomputation budgets or end-to-end cache-miss latency.

\end{document}